\documentclass[letterpaper, 10 pt, conference]{ieeeconf}  

\IEEEoverridecommandlockouts                              

\usepackage{graphics} 
\usepackage{times}

\usepackage{epsfig} 
\usepackage{graphicx}
\usepackage{cite}
\usepackage{makecell}
\usepackage{url}
\usepackage{amsmath} 
\usepackage{amssymb}  
\title{\LARGE \bf
Multi-Scale Feature Fusion with  Image-Driven Spatial Integration for Left Atrium Segmentation from Cardiac MRI Images}
\author{Bipasha Kundu$^{1}$, Zixin Yang$^{1}$, Richard Simon$^{2}$ and Cristian Linte$^{1,2}$
\thanks{$^{1}$Bipasha Kundu is with Center for Imaging Science, Rochester Institute of Technology
Rochester, NY 14623, USA.  {\tt\small bk7944@g.rit.edu}}%
\thanks{$^{1}$Zixin Yang is with Center for Imaging Science, Rochester Institute of Technology
Rochester, NY 14623, USA.  {\tt\small yy8898@g.rit.edu}}
\thanks{$^{2}$Richard Simon is with the Department of Biomedical Engineering, Rochester Institute of Technology, USA, NY
14623, USA.  {\tt\small rasbme@rit.edu}}
\thanks{$^{1,2}$Cristian Linte is with the Department of Biomedical Engineering, Rochester Institute of Technology, USA, NY
14623, USA. {\tt\small calbme@rit.edu}}
}
\begin{document}
\maketitle
\thispagestyle{empty}
\pagestyle{empty}
\begin{abstract}

Accurate segmentation of the left atrium (LA) from late gadolinium-enhanced magnetic resonance imaging plays a vital role in visualizing diseased atrial structures, enabling the diagnosis and management of cardiovascular diseases. It is particularly essential for planning treatment with ablation therapy, a key intervention for atrial fibrillation (AF). However, manual segmentation is time-intensive and prone to inter-observer variability, underscoring the need for automated solutions. Class-agnostic foundation models like DINOv2 have demonstrated remarkable feature extraction capabilities in vision tasks. However, their lack of domain specificity and task-specific adaptation can reduce spatial resolution during feature extraction, impacting the capture of fine anatomical detail in medical imaging. To address this limitation, we propose a segmentation framework that integrates DINOv2 as an encoder with a UNet-style decoder, incorporating multi-scale feature fusion and input image integration to enhance segmentation accuracy. The learnable weighting mechanism dynamically prioritizes hierarchical features from different encoder blocks of the foundation model, optimizing feature selection for task relevance.
Additionally, the input image is reintroduced during the decoding stage to preserve high-resolution spatial details, addressing limitations of downsampling in the encoder. We validate our approach on the LAScarQS 2022 dataset and demonstrate improved performance with a 92.3\% Dice and 84.1\% IoU score for giant architecture compared to the nnUNet baseline model. These findings emphasize the efficacy of our approach in advancing the field of automated left atrium segmentation from cardiac MRI.
\newline
\end{abstract}

\section{INTRODUCTION}
The segmentation of the left atrium (LA) from cardiac magnetic resonance (CMR) imaging plays a vital role in accurately assessing LA shape, size, and function, which are essential for diagnosing and managing various cardiovascular diseases. In particular, atrial fibrillation (AF), a prevalent arrhythmia in clinical practice, significantly benefits from precise imaging-based evaluation\cite{gonzales2021automated}. However, segmenting the LA from late gadolinium-enhanced (LGE) MRI remains challenging due to the inherent noise, low contrast between the atrial wall and surrounding structures, and variability in image quality across different patients and imaging protocols. Furthermore, the manual annotation process is time-consuming and inefficient, highlighting the need for an automated segmentation approach to efficiently process large volumes of CMR data \cite{tobon2014left}.

In recent years, Convolutional Neural Network (CNN) and Transformer-based segmentation methods have shown promising results for LA segmentation. Lefebvre \textit{et al.}\cite{lefebvre2022lassnet} used a region of interest (RoI) based two-stage neural network for LA segmentation. However, RoI-based methods are sensitive to region boundaries and downsample the resolution. Jiang \textit{et al.}\cite{jiang2022deep} proposed an end-to-end deep UNet architecture that strongly depends on extensive data augmentation to handle domain shifts. Additionally, CNN-based methods struggle with capturing long-range pixel relations. Hence, transformer based methods are capable of capturing long range dependencies, but are weak at capturing local variations\cite{wu2024transfusion},\cite{lin2024usformer}. In medical image segmentation, both the CNN and transformer-based networks face challenges in achieving accurate segmentation due to factors such as the morphological operations, varying anatomical structures of the organs, refined boundary, and limited annotated training data\cite{shi2024mlc}. 
\vspace{.8em}

The emergence of the vision foundation model (VFM) has introduced significant potential in the field of image segmentation. Among these, DINOv2 is considered the state-of-the-art (SOTA) for zero-shot segmentation in the natural domain. Several attempts have been proposed and tried to adapt DINOv2 for cross-domain generalization in different segmentation tasks\cite{xu2025generalizable}. Inspired by the strong segmentation capabilities of UNet, many researchers have explored it with a simple UNet decoder \cite{baharoon2023towards}. Although these general-purpose foundation models demonstrate strong capabilities in extracting consistent features across various domains, they often face challenges in capturing fine structural details. They process images in patch-based tokens and progressively convert abstract spatial details into low-resolution feature maps. While UNet decoders attempt to up-sample these features, fine structural details are lost during encoding, making pixel-level segmentation challenging. This limitation arises from the low-resolution nature of the representations generated by these models, which restricts their ability to delineate intricate spatial features effectively. Additionally, the presence of numerous transformer blocks in the encoder complicates the identification of which layers, shallow or deep, offer the most relevant feature information. 

To solve these challenges, we propose a segmentation framework that incorporates the pre-trained foundation model DINOv2 as encoders with multi-scale feature fusion that prioritizes the most contributive blocks across different depths and an input image integration to retain the high-resolution spatial details for LA segmentation along with UNet-style decoder. Using the rich representational capabilities of the foundation model combined with a robust and efficient decoder design, we explore the out-of-the-box potential of DINOv2 in the medical domain and achieve improved generalizability and segmentation accuracy compared to other state-of-the-art methods.

\section{Methods}

\begin{figure*} [t]
\centering
\includegraphics[width=0.8\linewidth]{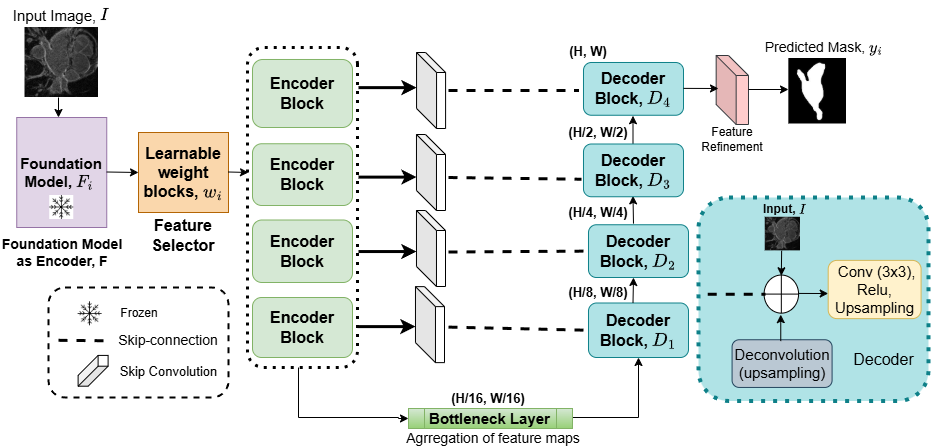}
\caption{Framework of the proposed UNet decoder-based segmentation model with learnable weight blocks, skip connections, input image augmentation, and decoder blocks}
  \label{fig1}
  \vspace{-.5em}
\end{figure*}

Our goal is to improve segmentation performance by effectively utilizing the features of the most relevant transformer blocks of the foundation model while incorporating spatial information and skip connections. Our previous study investigated foundation models with linear decoders, especially for reduced training data \cite{kundu2024assessing}. We further improve the segmentation performance by introducing a learnable-weight multi-level feature fusion strategy that dynamically assigns importance to each transformer block based on their contribution to the task, along with a UNet decoder to preserve and integrate the contextual information from each selected transformer block using a combination of dynamic skip connections, bottleneck layer, and input image integration. Fig. \ref{fig1} shows the overall workflow of our proposed segmentation framework.
\subsection{DINOv2 as Encoder}
DINOv2 is a vision transformer-based foundation model (ViT) developed and released by Meta \cite{oquab2023dinov2} in 2023. Our framework utilizes three DINOv2 architectures (DINOv2-giant, DINOv2-large, DINOv2-base) as the encoder to extract hierarchical features from the input images. The ViT-based architecture of DINOv2 first splits the input images $I \in \mathbb{R}^{H \times W \times C} $ into non-overlapping patches of size P $\times$ P (P = 14). These patches are linearly embedded in patch tokens, forming a sequence of patch embeddings $I \in \mathbb{R}^{N\times D}$, where $N =\frac{HW}{P^2}$ is the number of patches, D, H and W are the dimension, height, and weight of the embedding of the image, respectively. In our experiment, the DINOv2 encoder ($F$) was frozen during the training phase.

\subsection{Multi-scale feature fusion}
To dynamically determine the importance of each encoder block, we introduce a learnable block weight. Let $F_{i}$ denote the features from the $i$-th block, and $w_{i}$ represents the learned weight for that block. These weights are randomly initialized with small random values to ensure uniform importance at the start of training. During training, the optimization process adjusts these weights based on the loss gradients that allows the model to dynamically prioritize features based on their relevance to the task. The feature maps from all blocks are aggregated into a single tensor using the normalized weights following (1)
\[
\hat{F} = \sum_{i=1}^{N} w_i F_i \eqno{(1)}
\]
where N is the number of blocks in the foundation model encoder and $w_{i}$ are the weights, normalized using a softmax function in every forward pass used as in (2)
\[
w_i = \frac{\exp(\theta_i)}{\sum_{j=1}^{N} \exp(\theta_j)}  \eqno{(2)}
\]
where $\theta_{i}$ are the learnable parameters corresponding to each block. During training, these parameters $\theta_{i}$ are updated using back-propagation where the gradients are computed based on their contribution to the overall loss to allow the model to dynamically prioritize features based on their task-specific relevance. Blocks with higher weights contribute more, while the blocks with lower weights contribute less. We select the top k blocks based on the largest $w_{i} $ ensuring that decoder blocks process the most relevant features. This approach retains all encoder blocks, but dynamically adjusts their contributions, allowing the model to effectively utilize the hierarchical feature representations for task-specific adaptation and improved accuracy.
\subsection{Input Image Integration}
To enhance the accuracy and retain the spatial resolution of our segmentation mask, we incorporate the original input image into the decoding phases alongside features of the encoder coming through skip-connections and outputs from previous decoder blocks. The input image $I \in \mathbb{R}^{H \times W \times 3}$  is processed using a lightweight convolutional layer such that the transformed input $I'\in\mathbb{R}^{H \times W \times C}$ aligns with the decoder's feature dimensions. The original input image contains the complete spatial and pixel-level information (edges, textures, structures) lost during the downsampling operation of the encoder block. This additional pixel-level information from the input image helps the decoder preserve the fine-grained boundaries and global structures during the segmentation process. At each decoder stage, the concatenation of the transformed input image $I'$, foundation model encoder block $F_i$, and the output from the preceding decoder block $D_{j-1}$ is performed along the channel dimensions as described in (3).
\[
F_{concat} =  \text{concat}({I'}, {F}_{i}, {D}_{j-1})  \eqno{(3)}
\]
The resulting concatenated representation $F_{concat}$ is passed through the convolutional layers of the decoder block and the upsampling operations.
\subsection{UNet Decoder}
We designed a UNet-style decoder \cite{ronneberger2015u} to preserve the fine-grained high-resolution representation of the encoder block of the foundation model. After identifying the most important encoder blocks using learnable weights, the feature maps are reshaped to $ F \in \mathbb{R}^{B \times D \times N}$, where B is the batch size, N is the number of patches and D is the embedding dimension. We also used the last layer of the encoder block as the bottleneck layer. Unlike the classical UNet, we introduced a skip convolution to project the encoder features into a reduced dimension, ensuring compatibility with the corresponding decoder blocks. The decoder blocks are designed with a learnable upsampling layer followed by a convolutional layer, effectively doubling the spatial resolution at each stage. The resolution at each decoder block is given by $ \frac{H}{P}. {2^n}, \frac{W}{P}. {2^n}$, where n corresponds to the block level (n = 1x, 2x, 3x, 4x) from the bottom to the top of the decoder. The concatenated feature $F_{concat}$ is passed through a $3\times3$ convolutional layer to effectively integrate multi-scale features and refine spatial details for accurate segmentation for the further stage. Finally, another 3x3 convolution is applied to the final decoder ($D_{4}$) to refine and predict the final segmentation mask ($y_{i}$). 
\subsection{Dataset}
We trained and evaluated our approach on the LAScarQS 2022 dataset \cite{Lees-Miller-LaTeX-course-1}. We used the left atrium data from Task 1. The images were acquired using Siemens Avanto 1.5T, Philips Acheiva 1.5T, or Siemens Vario 3T scanners. The data set provides 60 labeled 3D LGE MRI images of resolution varying between 576 $\times$ 576 and 640 $\times$ 640 with a slice number of 44 or 88 in the third dimension. In our experiment, we split the data at the patient level and used 10\% of the training data for validation and 20\% for testing. 

\subsection{Implementation Details}
We used the PyTorch framework and conducted all the experiments in the RIT Research Computing Cluster \cite{https://doi.org/10.34788/0s3g-qd15} equipped with NVIDIA A100 40GB GPUs.  We utilized the Adam optimizer \cite{loshchilov2017decoupled} with two $\beta$ values ($\beta_{1}$= 0.90, $\beta_{2}$ = 0.95) and an initial learning rate of $5 \times 10^{-5}$ with a weight decay of $10^{-4}$. We also used a warm-up of 5 epochs and a Cosine Annealing learning rate scheduler. All the models were trained for 50 epochs with a batch size of 32 and with the default input image size of 448 $\times$ 448. We used the best validation checkpoints to evaluate our segmentation framework. 
\section{Results }
We evaluated the performance of our proposed method by calculating the mean Dice score and Intersection over Union (IoU) between the ground truth and the predicted segmentation masks. These metrics were used to compare our approach against the state-of-the-art method, nnUNet\cite{isensee2021nnu}, providing a comprehensive assessment of segmentation accuracy. The segmentation performance is summarized in Table \ref{tab1}. We observe that all DINOv2 models achieved higher scores than the baseline nnUNet model. The DINOv2-giant model performed with a Dice Score of 92.3\% and an IoU Score of 84.1\%, outperforming both DINOv2-base and DINOv2-large as well as the nnUNet model. Compared to nnUNet, DINOv2-giant demonstrated an improvement of 4.6\% in the Dice Score and 2.4\% in the IoU Score, highlighting its ability to leverage hierarchical features extracted by the vision transformer encoder. To compare the performance of the three DINOv2 model with our proposed modifications, we conducted a statistical (T-test) test and the DINOv2 model showed statistical significance in performance improvements with
p $<$ 0.05.
\begin{table}[t]
\caption{Segmentation evaluation, mean scores (\%) ± standard deviation for left atrium segmentation in LaScarQS dataset, statistical significance against nnUNet model represented by * for p $<$ 0.05}
\label{tab1}
\begin{center}

\begin{tabular}{|c|c|c|}

\hline
\textbf{Model Name} & \textbf{Mean Dice Score} & \textbf{Mean IoU Score} \\
\hline
\hline
nnUNet - 2D & 87.7 ± 2.7 & 81.7 ± 2.8 \\
\hline
DINOv2 - base & 91.5 ± 6.6 & 82.9 ± 9.5\\
\hline
DINOv2 - large & 92.1 ± 5.6 & 83.8 ± 8.4  \\
\hline
\textbf{DINOv2 - giant} &  \textbf{92.3 ± 5.9*} & \textbf{84.1 ± 8.8*}\\
\hline

\end{tabular}
\vspace{-0.8em}
\end{center}

\end{table}

\begin{figure} [!ht] 
\centering
\includegraphics[width=3.5in]{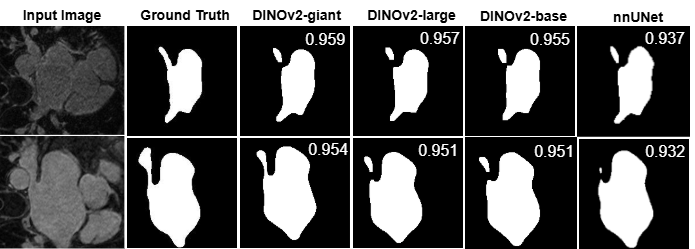}
\caption{Visualization of left atrium segmentation results comparing the baseline model and the proposed framework.}
\label{fig2}
\end{figure}

Fig. \ref{fig2} compares segmentation predictions from nnUNet and DINOv2 configurations (base, large, and giant) against the ground truth on two sample images. Dice scores are overlaid on each prediction. DINOv2-giant achieves the highest accuracy (e.g., 0.959), close to the ground truth, while nnUNet shows lower scores (e.g., 0.937) for the first example. This shows the improved performance of DINOv2 models for our proposed segmentation framework, particularly the giant configuration.

\section{Ablation Study}
We conducted an ablation study to evaluate the performance of our model using two configurations: utilizing the last four encoder blocks and the specific transformer blocks [7, 5, 3, 1] from the foundation model encoder with and without input image integration. Table \ref{tab2} presents the mean Dice scores achieved for each configuration. The results demonstrate that prioritizing specific encoder layers and including spatial integration significantly enhances the performance of our proposed framework for all configurations. At the same time, the absence of such prioritization leads to a noticeable decrease in segmentation accuracy.

\begin{table}[h]
\caption{Ablation study showcasing the Dice score (\%) for the selected encoder layers and their contributions to the segmentation performance in all models}
\label{tab2}
\begin{center}
\Huge
\resizebox{\columnwidth}{!}{%
\begin{tabular}{|c|c|c|c|c|}
\hline
\makecell{\textbf{Selected Blocks}\\ \textbf{from Encoder}} & \textbf{Spatial Integration} & \textbf{DINOv2-base} & \textbf{DINOv2-large} & \textbf{DINOv2-giant} \\

\hline
\hline
  Last 4  blocks& Yes & 88.1 ± 8.0 & 88.6 ± 6.5 & 89.8 ± 4.7  \\
  & No & 87.1 ± 6.4 & 86.6 ± 6.4 & 88.2 ± 4.8  \\
\hline
7, 5, 3, 1 & Yes & 88.3 ± 7.8  & 88.9 ± 5.8  &  90.3 ± 4.3\\
 & No & 87.7 ± 6.3  & 87.2 ± 6.6  &  88.5 ± 5.2\\
\hline
\end{tabular}
}
\vspace{-.8em}
\end{center}
\end{table}

\section{CONCLUSIONS}

This study presents a segmentation framework for the left atrium from cardiac MR images that combines the strengths of foundation models and task-specific architectural enhancements. By leveraging DINOv2 encoders with a learnable multi-scale feature fusion strategy, our framework dynamically selects and aggregates features from different encoder blocks based on their task relevance. The inclusion of image integration into the input during decoding preserves critical spatial and boundary details, addressing common limitations of traditional encoder-decoder architectures. From the evaluation, our method outperformed the state-of-the-art nnUNet baseline, achieving significant improvements in Dice and IoU scores. These findings underscore the ability of the proposed approach to provide accurate and reliable segmentation, allowing better clinical decision-making to manage AF and other cardiovascular conditions. 
\section*{ACKNOWLEDGMENT}
We would like to acknowledge the generous support for this work by the National Institutes of Health – National Institute of General Medical Sciences under Award No. R35GM128877 and the National Science Foundation - Division of Chemical, Bioengineering and Transport Systems under Award No. 2245152.


\begin{thebibliography}{10}

\bibitem{gonzales2021automated}
R.~A. Gonzales, F.~Seemann, J.~Lamy, P.~M. Arvidsson, E.~Heiberg, V.~Murray, and D.~C. Peters, ``Automated left atrial time-resolved segmentation in mri long-axis cine images using active contours,'' {\em BMC Medical Imaging}, vol.~21, no.~1, p.~101, 2021.

\bibitem{tobon2014left}
C.~Tobon-Gomez, J.~Peters, J.~Weese, K.~Pinto, R.~Karim, T.~Schaeffter, R.~Razavi, and K.~S. Rhode, ``Left atrial segmentation challenge: a unified benchmarking framework,'' in {\em Statistical Atlases and Computational Models of the Heart. Imaging and Modelling Challenges: 4th International Workshop, STACOM 2013, Held in Conjunction with MICCAI 2013, Nagoya, Japan, September 26, 2013. Revised Selected Papers 4}, pp.~1--13, Springer, 2014.

\bibitem{lefebvre2022lassnet}
A.~L. Lefebvre, C.~A. Yamamoto, J.~K. Shade, R.~P. Bradley, R.~A. Yu, R.~L. Ali, D.~M. Popescu, A.~Prakosa, E.~G. Kholmovski, and N.~A. Trayanova, ``Lassnet: A four steps deep neural network for left atrial segmentation and scar quantification,'' in {\em Challenge on Left Atrial and Scar Quantification and Segmentation}, pp.~1--15, Springer, 2022.

\bibitem{jiang2022deep}
L.~Jiang, Y.~Li, Y.~Wang, H.~Cui, Y.~Xia, and Y.~Zhang, ``Deep u-net architecture with curriculum learning for left atrial segmentation,'' in {\em Challenge on Left Atrial and Scar Quantification and Segmentation}, pp.~115--123, Springer, 2022.

\bibitem{wu2024transfusion}
M.~Wu, D.~Zhang, Y.~Hua, M.~Si, P.~Liu, and Q.~Wang, ``Transfusion: Efficient vision transformer based on 3d transesophageal echocardiography images for the left atrial appendage segmentation,'' {\em Expert Systems with Applications}, vol.~255, p.~124727, 2024.

\bibitem{lin2024usformer}
H.~Lin, S.~L{\'o}pez-Tapia, F.~Schiffers, Y.~Wu, S.~Gunasekaran, J.~Hwang, D.~Bishara, E.~Kholmovski, M.~Elbaz, R.~S. Passman, {\em et~al.}, ``Usformer: A small network for left atrium segmentation of 3d lge mri,'' {\em Heliyon}, vol.~10, no.~7, 2024.

\bibitem{shi2024mlc}
Z.~Shi, M.~Jiang, Y.~Li, B.~Wei, Z.~Wang, Y.~Wu, T.~Tan, and G.~Yang, ``Mlc: Multi-level consistency learning for semi-supervised left atrium segmentation,'' {\em Expert Systems with Applications}, vol.~244, p.~122903, 2024.

\bibitem{xu2025generalizable}
T.~Xu, S.~Hosseini, C.~Anderson, A.~Rinaldi, R.~G. Krishnan, A.~L. Martel, and M.~Goubran, ``A generalizable 3d framework and model for self-supervised learning in medical imaging,'' {\em arXiv preprint arXiv:2501.11755}, 2025.

\bibitem{baharoon2023towards}
M.~Baharoon, W.~Qureshi, J.~Ouyang, Y.~Xu, K.~Phol, A.~Aljouie, and W.~Peng, ``Towards general purpose vision foundation models for medical image analysis: An experimental study of dinov2 on radiology benchmarks,'' {\em arXiv preprint arXiv:2312.02366}, 2023.

\bibitem{kundu2024assessing}
B.~Kundu, B.~Khanal, R.~Simon, and C.~A. Linte, ``Assessing the performance of the dinov2 self-supervised learning vision transformer model for the segmentation of the left atrium from mri images,'' {\em arXiv preprint arXiv:2411.09598}, 2024.

\bibitem{oquab2023dinov2}
M.~Oquab, T.~Darcet, T.~Moutakanni, H.~Vo, M.~Szafraniec, V.~Khalidov, P.~Fernandez, D.~Haziza, F.~Massa, A.~El-Nouby, {\em et~al.}, ``Dinov2: Learning robust visual features without supervision,'' {\em arXiv preprint arXiv:2304.07193}, 2023.

\bibitem{ronneberger2015u}
O.~Ronneberger, P.~Fischer, and T.~Brox, ``U-net: Convolutional networks for biomedical image segmentation,'' in {\em Medical image computing and computer-assisted intervention--MICCAI 2015: 18th international conference, Munich, Germany, October 5-9, 2015, proceedings, part III 18}, pp.~234--241, Springer, 2015.

\bibitem{Lees-Miller-LaTeX-course-1}
LaScarQS, ``Left atrial and scar quantification \& segmentation challenge.'' 2022, \url{https://zmiclab.github.io/projects/lascarqs22}.
\newblock (Accessed: 20 January 2025).

\bibitem{https://doi.org/10.34788/0s3g-qd15}
R.~I. of~Technology, ``Research computing services,'' 2019.

\bibitem{loshchilov2017decoupled}
I.~Loshchilov, ``Decoupled weight decay regularization,'' {\em arXiv preprint arXiv:1711.05101}, 2017.

\bibitem{isensee2021nnu}
F.~Isensee, P.~F. Jaeger, S.~A. Kohl, J.~Petersen, and K.~H. Maier-Hein, ``nnu-net: a self-configuring method for deep learning-based biomedical image segmentation,'' {\em Nature methods}, vol.~18, no.~2, pp.~203--211, 2021.

\end{thebibliography}
\end{document}